%% file: arvix.tex
\documentclass[10pt,twocolumn,letterpaper]{article}

\usepackage[pagenumbers]{cvpr} 

\usepackage{graphicx}
\usepackage{amsmath}
\usepackage{amssymb}
\usepackage{booktabs}

\usepackage[pagebackref,breaklinks,colorlinks]{hyperref}

\usepackage[capitalize]{cleveref}
\crefname{section}{Sec.}{Secs.}
\Crefname{section}{Section}{Sections}
\Crefname{table}{Table}{Tables}
\crefname{table}{Tab.}{Tabs.}

\usepackage[linesnumbered,ruled]{algorithm2e}
\usepackage{multirow}
\usepackage{tabularx}
\usepackage{adjustbox}

\def\cvprPaperID{4895} 
\def\confName{CVPR}
\def\confYear{2022}

\addtocounter{footnote}{0}

\begin{document}

\title{PatchTrack: Multiple Object Tracking Using Frame Patches}

\author{Xiaotong Chen\thanks{The work is done during an internship at Appen}\\
Computer Science\\
UC, Santa Barbara\\
{\tt\small xchen774@ucsb.edu}
\and
Seyed Mehdi Iranmanesh\thanks{The work is done while at Appen}\\
Amazon\\
{\tt\small  mehdiir@amazon.com}
\and
Kuo-Chin Lien\\
Appen\\
{\tt\small  klien@appen.com}
}

\maketitle

\input{sections/Abstract}
\input{sections/Introduction}

\input{sections/RelatedWork}
\input{sections/Method}

\input{sections/Experiments}

\input{sections/Conclusion}

{\small
\bibliographystyle{ieee_fullname}
\bibliography{egbib}
}

\end{document}

%% file: sections/Abstract.tex
\begin{abstract}
Object motion and object appearance are commonly used information in multiple object tracking (MOT) applications, either for associating detections across frames in tracking-by-detection methods or direct track predictions for joint-detection-and-tracking methods. However, not only are these two types of information often considered separately, but also they do not help optimize the usage of visual information from the current frame of interest directly. In this paper, we present PatchTrack, a Transformer-based joint-detection-and-tracking system that predicts tracks using patches of the current frame of interest. We use the Kalman filter to predict the locations of existing tracks in the current frame from the previous frame. Patches cropped from the predicted bounding boxes are sent to the Transformer decoder to infer new tracks. By utilizing both object motion and object appearance information encoded in patches, the proposed method pays more attention to where new tracks are more likely to occur. We show the effectiveness of PatchTrack on recent MOT benchmarks, including MOT16 (MOTA 73.71\%, IDF1 65.77\%) and MOT17 (MOTA 73.59\%, IDF1 65.23\%). The results are published on \href{{https://motchallenge.net/method/MOT=4725&chl=10}}{https://motchallenge.net/method/MOT=4725\&chl=10}.
\end{abstract}

%% file: sections/Introduction.tex
\begin{figure}
\begin{subfigure}[b]{\columnwidth}
\centering
\includegraphics[width=0.55\columnwidth]{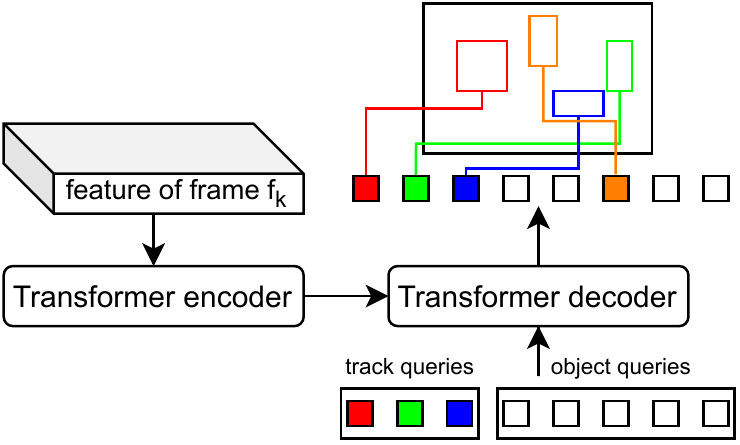}
\caption{Tracking using tracking queries (output embeddings of the previous frame).}
\label{fig:introduction_trackformer}
\end{subfigure}
\vspace{0.03cm}

\begin{subfigure}[b]{\columnwidth}
\centering
\includegraphics[width=0.55\columnwidth]{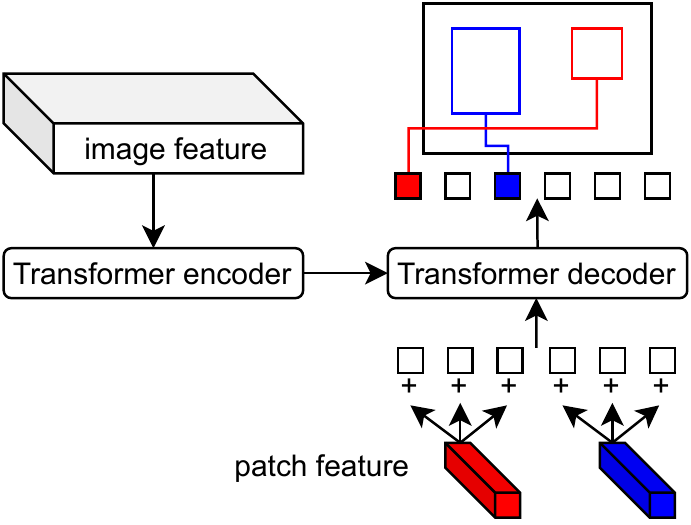}
\caption{Detection pre-trained to detect image patches.}
\label{fig:introduction_updetr}
\end{subfigure}
\vspace{0.03cm}

\begin{subfigure}[b]{\columnwidth}
\centering
\includegraphics[width=0.9\columnwidth]{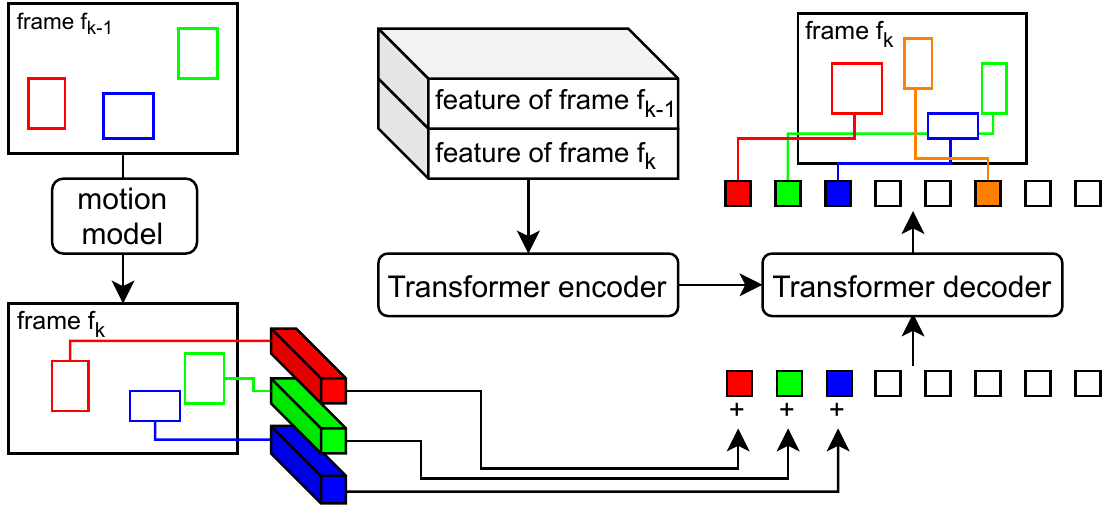}
\caption{Tracking using patch queries and frame patches.}
\label{fig:introduction_patchtrack}
\end{subfigure}
\caption{\textbf{Inspiration of PatchTrack.} MOT method~\cite{sun2020transtrack, meinhardt2021trackformer} uses output embeddings of the previous frame as track queries (colors represent tracking IDs) to propagate existing objects and object queries to detect new objects entering the camera view (\ref{fig:introduction_trackformer}). Detection model~\cite{dai2021up} is pre-trained to locate image patches by adding their features to object queries (\ref{fig:introduction_updetr}). Proposed system that uses both track queries and features of frame patches predicted by a motion model to predict locations of corresponding objects (\ref{fig:introduction_patchtrack}).}
\label{fig:introduction}
\end{figure}

\section{Introduction}

Multiple object tracking (MOT) concerns identifying objects of interest and tracking their moving trajectories in video sequences. Intuitively, successful MOT algorithms need to be able to handle subtle appearance differences between multiple tracked objects and resolve the ambiguity via other cues, such as motion, when the targets are visually indistinguishable. 

With the powerful appearance encoding capability of CNN, the tracking-by-detection paradigm dominates MOT methods in the past decade~\cite{carion2020end, zhang2020fairmot, Wojke2017simple}. Highly accurate CNN-based object detection~\cite{redmon2016you, ren2015faster, cai2018cascade} is first performed in all frames independently, and then association of these detected objects across frames is performed to establish tracks of consistent object IDs. In the association step, locations of existing tracks in the following frame may be predicted from assumption (constant velocity, acceleration, etc.) or other motion models~\cite{zhang2020fairmot, shuai2021siammot, welch1995introduction, Wojke2017simple} and then associate with detections based on metrics like intersection-over-union (IoU).

Joint-detection-and-tracking methods~\cite{zhou2020tracking, sun2020transtrack, wu2021track} recently demonstrate superior accuracy. The idea is simultaneously performing object detection and tracking so both tasks enjoy information shared from the other. This is particularly intriguing in Transformer-based architectures where output feature embeddings of previous frames are used as ‘track queries’, along with ‘object queries’ for Transformer decoder, predicting corresponding tracks as well as newly discovered objects in the current frame (Figure \ref{fig:introduction_trackformer}). Albeit achieving state of the art MOT results, we argue that these architectures overly rely on appearance. As the information encoded in track queries is strictly limited to previous frames, the Transformer model needs to infer both object offset and object appearance in the current frame.

To resolve above problem, we take inspiration from UP-DETR~\cite{dai2021up}, an object detection model that is pre-trained to detect image patches (Figure \ref{fig:introduction_updetr}) using patch features, and propose a MOT system that uses frame patches from the current frame of interest. We first use a motion model to predict new locations of existing tracks in the current frame from the previous frame, and crop the current frame to patches based on the prediction. These patches, with implicit prior knowledge of object motion and explicit information of object appearance in the current frame, are sent to the decoder to predict new locations of existing tracks in the current frame.  

More specifically, we present PatchTrack (Figure \ref{fig:introduction_patchtrack}), which is a Transformer-based joint-object-detection-and-tracking system that predicts tracks in the current frame of interest from its patches. We use the Kalman filter~\cite{welch1995introduction} to obtain track candidates in the current frame from existing tracks in the previous frame and crop the current frame using the bounding box of these candidates to get patches. Both the current frame and these patches are sent into our convolutional neural network (CNN)~\cite{Goodfellow-et-al-2016} backbone that outputs the frame feature and the patch queries respectively. Each pair of track query, from the output embeddings when processing the previous frame, and patch query with the same tracking ID are added together to form the corresponding patch-track query. These patch-track queries are sent to the decoder along with object queries, where the former is used to predict new locations of existing tracks, while the latter is used to detect new objects in the current frame.

We evaluate PatchTrack on MOT benchmarks and achieve competitive results on MOT16 (MOTA 73.71\%, IDF1 65.77\%) and MOT17 (MOTA 73.59\%, IDF1 65.23\%) test sets. To the best of our knowledge, our method is the first that uses patches of the current frame of interest to infer both object motion and appearance information simultaneously.  We hope it could provide a new perspective for designing MOT systems. 

In summary, our contributions are:
\begin{itemize}
\item A Transformer-based MOT system, namely PatchTrack, which jointly performs object detection and tracking.
\item A novel way of optimizing the usage of visual information by utilizing patches from the current frame of interest. 
\item Introduction of patch-track queries that incorporate both knowledge of the object motion and object appearance in the current frame of interest to facilitate tracking. 
\end{itemize}

%% file: sections/RelatedWork.tex
\section{Related Work}

\subsection{Object detection and tracking}
Object detection concerns locating and/or classifying objects of interest in a single image. As the preliminary to object tracking, there is a close connection between the two. Many popular object detection methods generate detections from hypothesis of object locations, including regional proposals~\cite{girshick2014rich, girshick2015fast, ren2015faster, cai2018cascade} and anchors/object centers~\cite{redmon2016you, liu2016ssd, zhou2019objects}. On the other hand, there is an increasing number of object tracking systems that utilize Transformer~\cite{vaswani2017attention}, which has shown success in object detection~\cite{carion2020end, meng2021conditional, zhu2020deformable, dai2021up} before. Transformer-based object-detection methods encode the CNN~\cite{Goodfellow-et-al-2016} feature of images and decodes learned object queries to obtain detections. Aside from architecture adjustment~\cite{meng2021conditional, zhu2020deformable} from the original DETR~\cite{carion2020end}, we also see modification to object queries~\cite{dai2021up} using image patches to facilitate detection. Inspired by the usage of regional proposal and image patches, our proposed method uses frame patches, which can be considered as our initial guess of track locations and appearance.

\subsection{Tracking-by-detection}
One major paradigm in MOT is tracking-by-detection, where the MOT systems~\cite{carion2020end, zhang2020fairmot, Wojke2017simple} first obtain detections for each frame and then associate them across frames to form tracks. Since the object detection is a standalone step in the tracking process, one benefit of tracking-by-detection methods is the flexibility to pair different object detection models~\cite{ren2015faster, redmon2016you, carion2020end} with different association strategies, thus be benefited directly from advancement in the area of object detection. On the other hand, the object detection step omits information across frames as each of them is processed separately by the detector. 

Object motion and appearance may only be considered as part of the detection association strategy for these methods~\cite{zhang2020fairmot, shuai2021siammot}. For object motion, Kalman filter~\cite{welch1995introduction} is one of the most popular algorithm used to propagate detections in previous frame to predict their location in its future frame. Combined with Hungarian algorithm~\cite{kenesei2002hungarian} and intersection-over-union (IOU) metrics, it has proven to be an effective tracking mechanism~\cite{bewley2016simple}. Object appearance information like Re-ID features~\cite{Wojke2017simple, pang2021quasi, zhang2020fairmot} are also commonly used as similarity measures.

\subsection{Joint-detection-and-tracking}
The other popular paradigm in MOT is joint-detection-and-tracking, where the object detection and object tracking are performed simultaneously~\cite{zhou2020tracking, sun2020transtrack}. One advantage of joint-detection-and-tracking methods is the accessibility to information across frames. For instance, features of multiple frames can be used at once~\cite{zhou2020tracking, sun2020transtrack, wu2021track} to facilitate detection and/or tracking. For Transformer-based joint-detection-and-tracking methods, both the encoder and the decoder may take additional information from previous frames to infer predictions of the current frame of interest~\cite{sun2020transtrack, meinhardt2021trackformer, zeng2021motr}. Specifically, recent works have introduced track queries~\cite{sun2020transtrack, meinhardt2021trackformer}, which come from the output embeddings when processing previous frames. Depends on the design, the track queries may be decoded to bounding boxes separately from the object queries~\cite{sun2020transtrack} and matched together to predict new tracks, or processed together to form new tracks directly~\cite{meinhardt2021trackformer}.

%% file: sections/Method.tex
\begin{figure*}
\centering
\includegraphics[width=\textwidth]{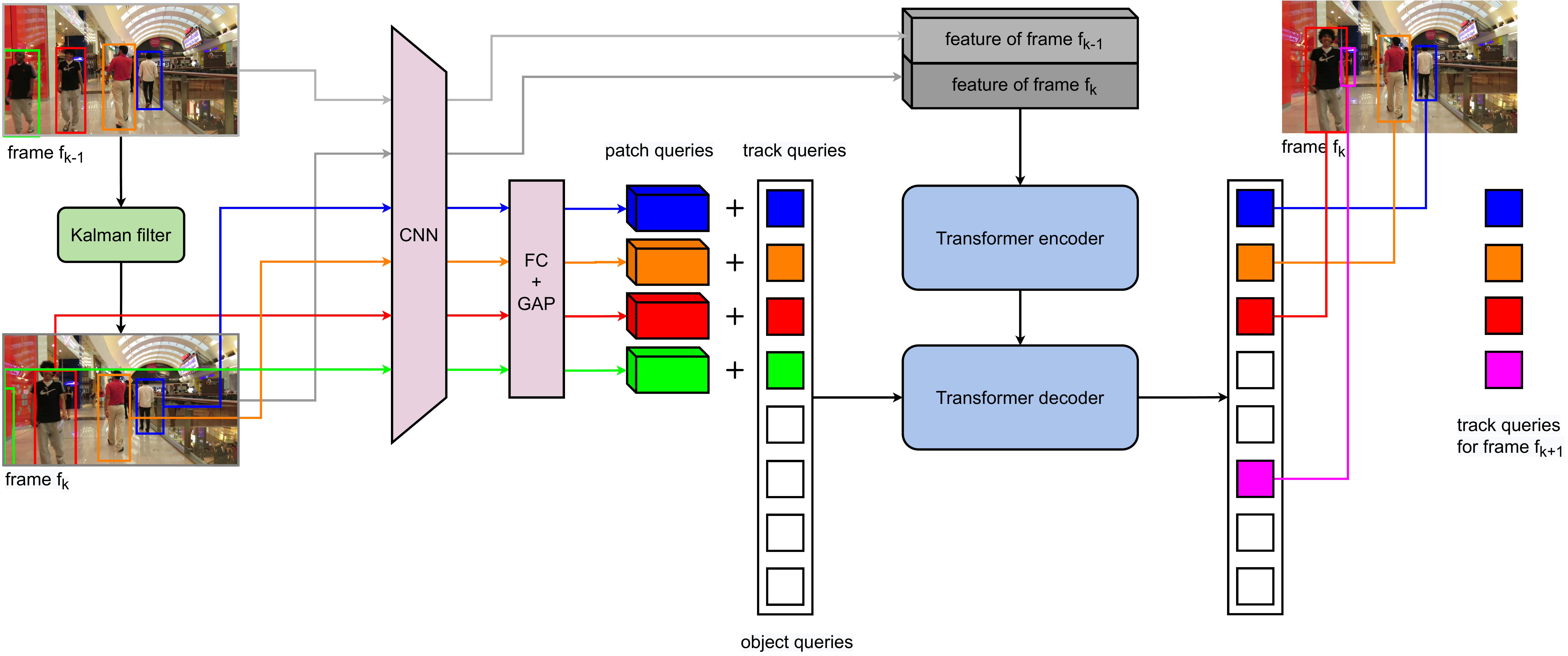}
\caption{\textbf{PatchTrack.} We first use Kalman filter~\cite{welch1995introduction} to predict track candidates in frame $f_{k}$ from tracks in frame $f_{k-1}$. Both frames are sent to the CNN backbone that produces frame features for the Transformer encoder. We crop $f_k$ to patches using bounding boxes of track candidates and send them to the CNN backbone, followed by a fully connected layer (FC) and global average pooling (GAP), to get patch queries that align with track queries. Patch queries are added to track queries to form patch-track queries, which are then sent to the Transformer decoder along with object queries. The patch-track queries are decoded to output embeddings that refine locations of track candidates and the object queries are decoded to output embeddings that detect new objects. Output embeddings that corresponds to tracks in $f_k$ are the track queries for processing $f_{k+1}$. }
\label{fig:method_patchtrack}
\end{figure*}

\section{Method}

In this section, we describe the architecture of PatchTrack (Section~\ref{section:architecture}), how object tracking is initialized (Section~\ref{section:object_tracking_initalization}), how existing tracks are propagated to form new track candidates (Section~\ref{section:track_propagation}), and how frame patches are generated to help facilitate object tracking(Section \ref{section:patch_generation_and_object_tracking}). 

\subsection{Architecture\label{section:architecture}}

PatchTrack is a Transformer-based joint-detection-and-tracking system. The Transformer encoder takes in CNN features of a consecutive frame pair. The Transformer decoder takes queries as input and output bounding boxes. PatchTrack deals with four types of queries: object queries, track queries, patch queries, and patch-track queries. Depending on the source of the queries, the predicted bounding boxes may correspond to either tracks associated with existing tracking IDs or detections that need to be assigned with new tracking IDs.

\subsection{Object tracking initialization \label{section:object_tracking_initalization}}

Object tracking for the first frame $f_1$ is equivalent to object detection, where each predicted detection can be arbitrarily assigned to a unique tracking ID to form tracks. Frame $f_1$ is sent to the CNN backbone that outputs the corresponding frame feature. This feature is stacked with itself~\cite{sun2020transtrack} and sent to the Transformer encoder. Since there are no existing tracks to form non-object queries, the Transformer decoder only takes object queries as input and produces embeddings. The output embeddings that result in the non-background bounding boxes are the predicted detection in $f_1$, each of which is assigned to a unique tracking ID to form tracks. These embeddings are also used as the track queries for the next frame.

\subsection{Track propagation\label{section:track_propagation}}

For frame $f_k$ ($k > 1$), there exists $f_{k-1}$ with a set of tracks $T_{k-1}$. We can propagate these tracks using a motion model and infer tracks in $f_k$ (Algorithm \ref{alg:object_propogation}). 

Here we use the Kalman filter~\cite{welch1995introduction} as our motion model to predict a set of track candidates for $f_k$, namely $\widehat{T}_k$. The reason we call them track \textit{candidates} is because there are several problems if we use them directly as tracks in $f_k$. First of all, since the tracks in $\widehat{T}_k$ are mapped one-to-one with the ones in $T_{k-1}$, they only include objects that have appeared in $f_{k-1}$. Secondly, although Kalman filter and other motion models have shown effectiveness in many cases~\cite{bewley2016simple, veeramani2018deepsort, zhang2020fairmot}, their predicted bounding boxes are not accurate enough in terms of locating objects. This is the reason why motion models are often used to process existing tracks, and IoU is introduced to match processed tracks with new detections to form new tracks. In the paradigm of joint-object-detection-and-tracking, our architecture is designed to refine these track candidates to more accurate tracks.

\begin{algorithm}
    \SetKwInOut{Input}{Input}
    \SetKwInOut{Output}{Output}

    \Input{Tracks $T_{k-1}$ in frame $f_{k-1}$;\\ Motion model \texttt{M}}
    \Output{Track candidates $\widehat{T}_{k}$ for frame $f_k$}
    Initialization: $\widehat{T}_k \leftarrow \emptyset$\;
    \For{$t \in T_{k-1}$}
      {
        $\widehat{T}_k \leftarrow \widehat{T}_k \cup \{\texttt{M}(t)\}$ \;
      }
    \caption{Pseudo-code for object propagation}
    \label{alg:object_propogation}
\end{algorithm}

\subsection{Patch generation and object tracking \label{section:patch_generation_and_object_tracking}}

To tackle the above problems, we take inspiration from UP-DETR~\cite{dai2021up} where its Transformer decoder is pre-trained to detect locations of random image patches using their corresponding CNN features. Our proposed PatchTrack takes patches of frame $f_k$ as additional visual information besides the entire $f_k$ to perform object tracking. Specifically, for each track candidate $\widehat{t} \in \widehat{T}_k$, we crop the frame using its bounding box and send the resulting patch to the CNN backbone to get the corresponding patch feature. We use a fully-connected (FC) layer followed by global average pooling (GAP) to process all patch features to patch queries that align with track queries (Figure \ref{fig:method_patchtrack}). Each patch query is added to the track query from the same tracking ID to form a patch-track query. The patch-track queries are sent to the Transformer decoder alone with the initial object queries, both of which are processed jointly. Output embedding decoded from each patch-track query may either correspond to the refined location of the corresponding track candidate, or the background if the object has left $f_k$. On the other hand, the embeddings decoded from object queries that result in non-background detections locate new objects entering $f_{k}$, which are assigned with new tracking IDs to form new tracks. All embeddings that contribute tracks in $f_k$ form the track queries for $f_{k+1}$ (Figure \ref{fig:method_patchtrack}).  

\subsection{Track re-birth}

To obtain track queries for frame $f_{k+1}$ from the track queries for $f_{k}$, embeddings corresponding to the new detections are added and track queries corresponding to background class are removed (Figure \ref{fig:method_patchtrack}). A problem with this mechanism is that it is not robust to long-range tracking: if one object is not successfully detected, it can only be assigned to a new tracking ID when it is detected again, which causes fragmented trajectories. To tackle this problem, we adopt the track re-identification strategy from TrackFormer~\cite{meinhardt2021trackformer} and store these originally removed patch-track queries to an inactive query set. Queries in this set are included in the list of patch-track queries and sent to the decoder for at most $P$ consecutive frames. If the queries can be decoded to non-background bounding boxes during this process, these queries are re-activated with their original tracking IDs, otherwise they will be removed.

\subsection{Set prediction loss}

As shown in the model architecture~\ref{fig:method_patchtrack}, PatchTrack processes a frame pair $f_{k-1}$ and $f_{k}$ iteratively, and there are two steps involved. The first step is performing object detection on $f_{k-1}$ in order to initialize track queries for processing $f_k$ later. The second step is to perform object tracking on $f_k$ using previously generated track queries. Since the second steps involves detecting new objects, which is the same as the first step, as well as tracking existing object with tracking IDs associated with track queries, we use two set prediction loss~\cite{carion2020end}, one for detection new objects and the other for tracking objects existing in $f_{k-1}$. 

Let us denote $T_{k-1}$ and $T_k$ as the tracks for $f_{k-1}$ and $f_k$ respectively. In the case of detecting new objects, we are looking at any track $t \in T_k \setminus T_{k-1}$, which corresponds to new objects in $f_k$ but not $f_{k-1}$. We adopt object detection set prediction loss following the matching cost in TransTrack~\cite{sun2020transtrack} and DETR~\cite{carion2020end}:
\begin{equation}
\mathcal{L}_{det} = \lambda_{cls} \cdot \mathcal{L}_{det\_cls} + \lambda_{L1} \cdot \mathcal{L}_{det\_L1} + \lambda_{IoU} \cdot \mathcal{L}_{det\_IoU},
\end{equation}
where $\mathcal{L}_{det\_cls}$ is the focal loss~\cite{lin2017focal} between predicted class labels and the ground truth, $\mathcal{L}_{det\_L1}$ and $\mathcal{L}_{det\_IoU}$ are L1 loss and generalized IoU loss~\cite{rezatofighi2019generalized} between the normalized center and sides of the predicted bounding boxes and ground truth, while $\lambda_{cls}$, $\lambda_{L1}$ and $\lambda_{IoU}$ are their weights respectively. Predictions generated from decoding object queries are compared with the ground truth $t \in T_k \setminus T_{k-1}$, so $\mathcal{L}_{det}$ handles new object detection. 

Similarly, our object tracking set prediction loss is as follows:
\begin{equation}
\mathcal{L}_{trk} = \lambda_{cls} \cdot \mathcal{L}_{trk\_cls} + \lambda_{L1} \cdot \mathcal{L}_{trk\_L1} + \lambda_{IoU} \cdot \mathcal{L}_{trk\_IoU},
\end{equation}
where $\mathcal{L}_{trk\_cls}$, $\mathcal{L}_{trk\_L1}$, and $\mathcal{L}_{trk\_IoU}$ are calculated between predictions generated from decoding patch-track queries and the ground truth $t \in T_k \cap T_{k-1}$, so $\mathcal{L}_{trk}$ handles tracking objects in $f_{k-1}$ and predict their new locations in $f_k$. 

Our final loss function is simply the sum of object detection set prediction loss and object tracking set prediction loss: $\mathcal{L} = \mathcal{L}_{det} + \mathcal{L}_{trk}$.

%% file: sections/Experiments.tex
\newcolumntype{L}{>{\hsize=2\hsize}X}
\newcolumntype{M}{X}
\newcolumntype{S}{>{\hsize=.5\hsize}X}

\begin{table*}[]
\centering
\begin{tabularx}{\textwidth}{S|M|SSSSSSS}
\hline
Dataset                 & Method        & MOTA$\uparrow$ & IDF1$\uparrow$ & MT$\uparrow$   & ML$\downarrow$   & FP$\downarrow$     & FN$\downarrow$     & IDsw$\downarrow$  \\ \hline
\multirow{16}{*}{MOT16} & DeepSORT~\cite{Wojke2017simple} & 61.4 & 62.2 & 32.8 & 18.2 & 12,852 & 56,668 & 781   \\
                        & HTA~\cite{lin2021detection}           & 62.4 & 64.2 & 37.5 & 12.1 & 19,071 & 47,839 & 1,619 \\
                        & VMaxx~\cite{wan2018multi}         & 62.6 & 49.2 & 32.7 & 21.1 & 10,604 & 56,182 & 1,389 \\
                        & RAR16~\cite{fang2018recurrent}         & 63.0 & 63.8 & 39.9 & 22.1 & 13,663 & 53,248 & 482   \\
                        & TAP~\cite{zhou2018online}           & 64.8 & \textbf{73.5} & 40.6 & 22.0 & 12,980 & 50,635 & 794   \\
                        & CNNMTT~\cite{mahmoudi2019multi}        & 65.2 & 62.2 & 32.4 & 21.3 & 6,578  & 55,896 & 946   \\
                        & POI~\cite{yu2016poi}           & 66.1 & 65.1 & 34.0 & 21.3 & \textbf{5,061}  & 55,914 & 805   \\
                        & GSDT~\cite{wang2021joint}          & 66.7 & 69.2 & 38.6 & 19.0 & 14,754 & 45,057 & 959   \\
                        & TubeTK~\cite{pang2020tubetk}        & 66.9 & 62.2 & 39.0 & 18.1 & 11,544 & 47,502 & 1,236 \\
                        & LM\_CNN~\cite{babaee2019dual}       & 67.4 & 61.2 & 38.2 & 19.2 & 10,109 & 48,435 & 931   \\
                        & Chain-Tracker~\cite{peng2020chained} & 67.6 & 57.2 & 32.9 & 23.1 & 8,934  & 48,305 & 1,897 \\
                        & KDNT(POI)~\cite{yu2016poi}          & 68.2 & 60.0 & 41.0 & 19.0 & 11,479 & 45,605 & 933   \\
                        & FairMOT~\cite{zhang2020fairmot}          & 69.3 & 72.3 & 40.3 & 16.7 & 13,501 & 41,653 & 815   \\
                        & QuasiDense~\cite{pang2021quasi}    & 69.8 & 67.1 & 41.6 & 19.8 & 9,861  & 44,050 & 1,097 \\
                        & TraDeS~\cite{wu2021track}        & 70.1 & 64.7 & 37.3 & 20.0 & 8,091  & 45,210 & 1,144 \\
                        & LMP\_p~\cite{tang2017multiple}           & 71.0 & 70.1 & \textbf{46.9} & 21.9 & 7,880  & 44,564 & \textbf{434}  \\
                        & \textbf{PatchTrack (Ours)} & \textbf{73.3} & 65.8 & 45.7 & \textbf{11.3} & 10,660 & \textbf{36,824} & 1,179 \\ \hline
\end{tabularx}
\caption{\textbf{Evaluation on the MOT16 test set.} We evaluate recent MOT systems on the MOT16 test set in the private detection protocol. The method names are taken directly from the leaderboard of  \href{https://motchallenge.net/}{motchallenge}, where the names in parentheses are associated with their respective literatures. Metrics with $\uparrow$ means higher numbers are preferable, while the ones with  $\downarrow$ means lower numbers are preferable. Numbers are marked in bold if they are the best in their respective metric columns. Our proposed PatchTrack achieves best results in MOTA, ML, and FN.}
\label{tab:experiments_mot16}
\end{table*}

\begin{figure*}
\centering
\begin{subfigure}[b]{0.3\textwidth}
\centering
\includegraphics[width=\textwidth]{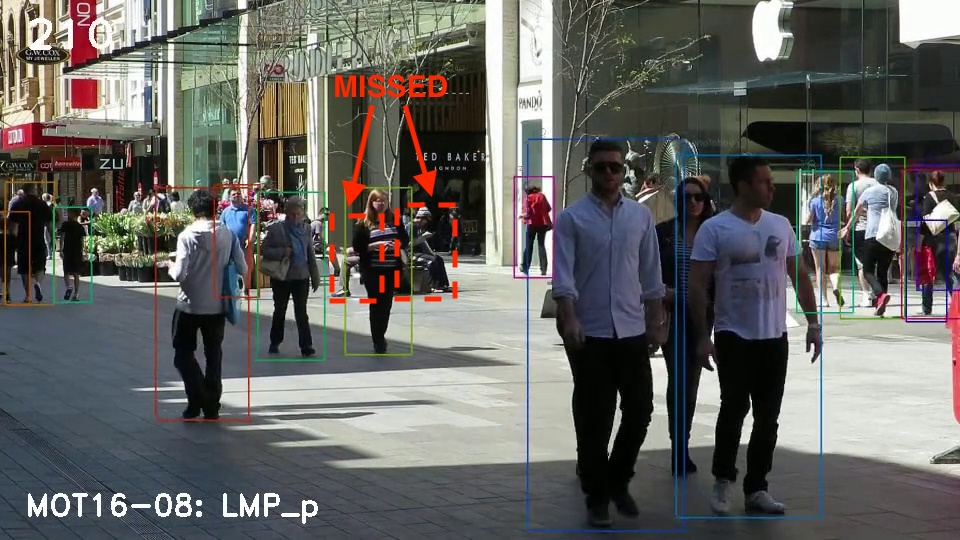}
\caption{LMP\_p MOT16-08 Frame 210}
\label{fig:experiments_lmp_mot16_08_210}
\end{subfigure}
\begin{subfigure}[b]{0.3\textwidth}
\centering
\includegraphics[width=\textwidth]{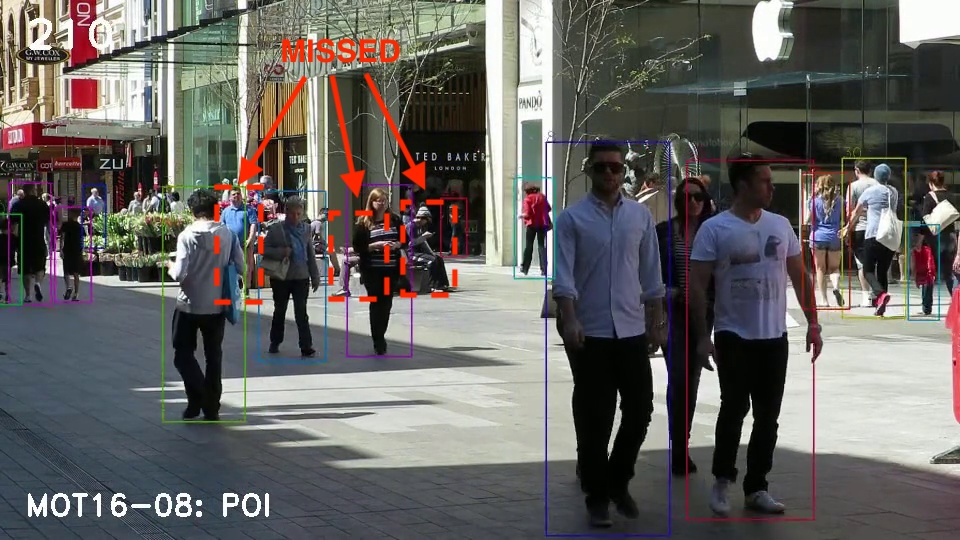}
\caption{POI MOT16-08 Frame 210}
\label{fig:experiments_poi_mot16_08_210}
\end{subfigure}
\begin{subfigure}[b]{0.3\textwidth}
\centering
\includegraphics[width=\textwidth]{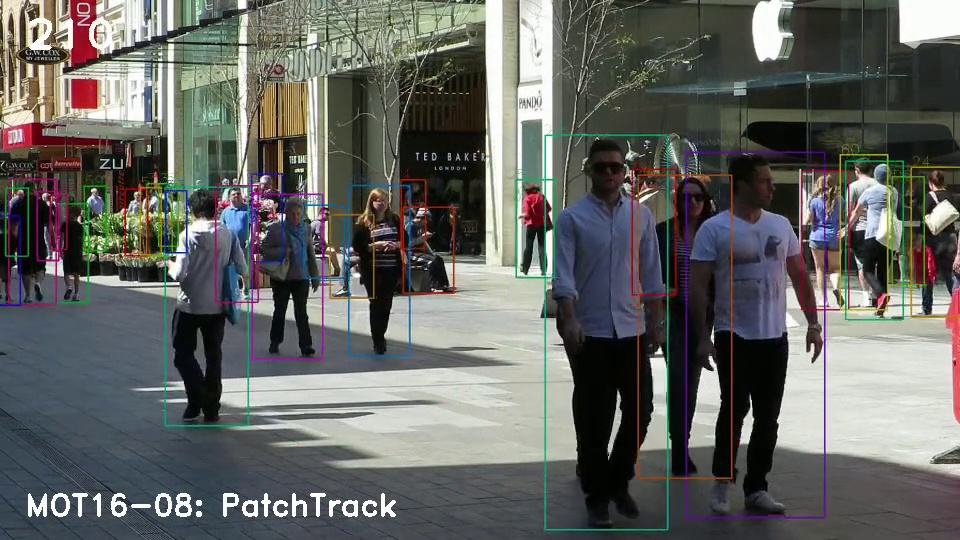}
\caption{PatchTrack MOT16-08 Frame 210}
\label{fig:experiments_patchtrack_mot16_08_210}
\end{subfigure}

\begin{subfigure}[b]{0.3\textwidth}
\centering
\includegraphics[width=\textwidth]{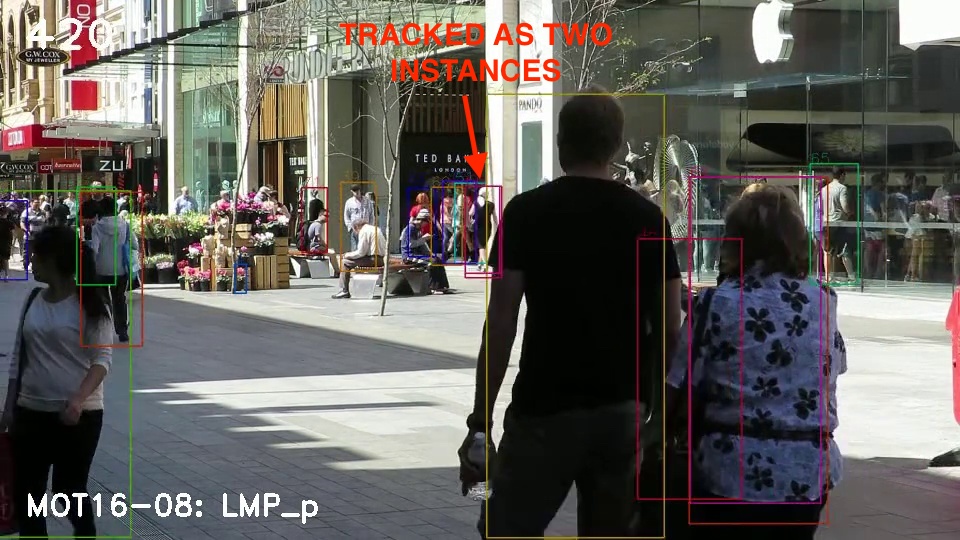}
\caption{LMP\_p MOT16-08 Frame 420}
\label{fig:experiments_lmp_mot16_08_420}
\end{subfigure}
\begin{subfigure}[b]{0.3\textwidth}
\centering
\includegraphics[width=\textwidth]{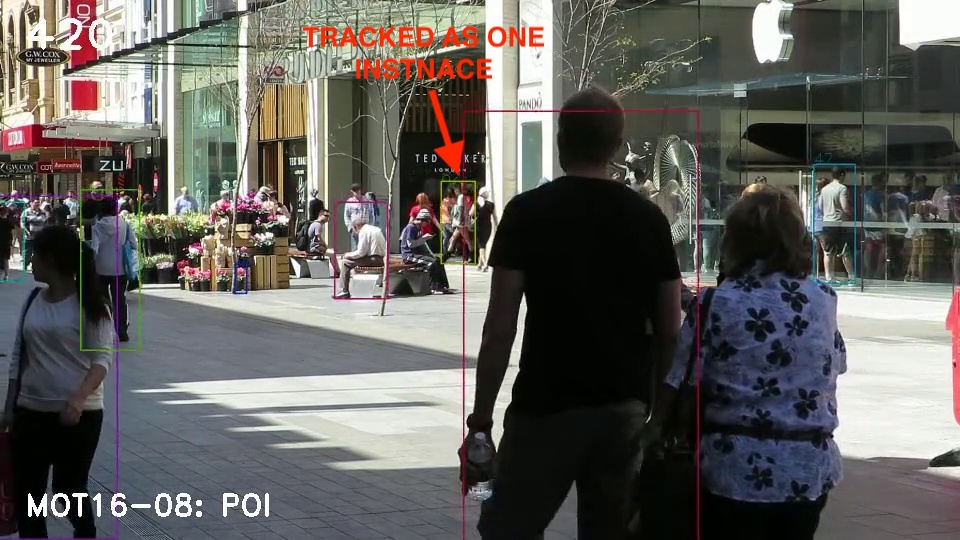}
\caption{POI MOT16-08 Frame 420}
\label{fig:experiments_poi_mot16_08_420}
\end{subfigure}
\begin{subfigure}[b]{0.3\textwidth}
\centering
\includegraphics[width=\textwidth]{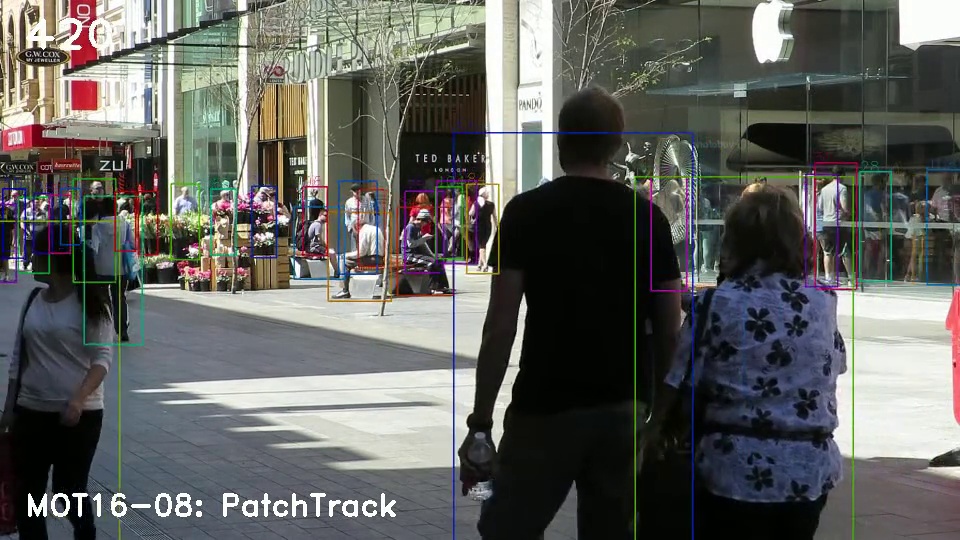}
\caption{PatchTrack MOT16-08 Frame 420}
\label{fig:experiments_patchtrack_mot16_08_420}
\end{subfigure}
\caption{\textbf{Visualizations on the MOT16 test set.} Visualizations on the MOT16 test set are taken from \href{https://motchallenge.net/}{motchallenge}. We add additional annotations in red to show challenging cases where LMP\_p~\cite{tang2017multiple} and POI~\cite{yu2016poi} fail to track. While both LMP\_p (Figure~\ref{fig:experiments_lmp_mot16_08_210}) and POI (Figure~\ref{fig:experiments_poi_mot16_08_210}) fail to track objects that are partially occluded, PatchTrack is still able to locate such objects (Figure~\ref{fig:experiments_patchtrack_mot16_08_210}). Additionally, PatchTrack performs better in distinguish different objects in a cluster (Figure~\ref{fig:experiments_patchtrack_mot16_08_420}) without missing (Figure~\ref{fig:experiments_poi_mot16_08_420}) objects or tracking one object twice (Figure~\ref{fig:experiments_lmp_mot16_08_420}).
}
\label{fig:experiments_mot16_visualization}
\end{figure*}

\begin{table*}[]
\centering
\begin{tabularx}{\textwidth}{S|L|SSSSSSS}
\hline
Dataset                 & (CNN-based) method                   & MOTA$\uparrow$                 & IDF1$\uparrow$                 & MT$\uparrow$                   & ML$\downarrow$                   & FP$\downarrow$                   & FN$\downarrow$                   & IDsw$\downarrow$                 \\ \hline
\multirow{23}{*}{MOT17} & DAN~\cite{sun2019deep}                     & 52.4                 & 49.5                 & 21.4                 & 30.7                 & 25,423               & 234,592              & 8,431                \\
                        & TubeTK~\cite{pang2020tubetk}                   & 63.0                 & 58.6                 & 31.2                 & 19.9                 & 27,060               & 177,483              & 4,137                \\
                        & GSDT~\cite{wang2021joint}                     & 66.2                 & 63.4                 & 36.9                 & 21.7                 & 25,800               & 164,120              & 2,711                \\
                        & Chained-Tracker~\cite{peng2020chained}          & 66.6                 & 57.4                 & 37.8                 & 18.5                 & 22,284               & 160,491              & 5,529                \\
                        & CenterTrack~\cite{zhou2020tracking}              & 67.8                 & 64.7                 & 34.6                 & 24.6                 & \textbf{18,498}      & 160,332              & 3,039                \\
                        & QuasiDense~\cite{pang2021quasi}              & 68.7                 & 66.3                 & 40.6                 & 21.9                 & 26,589               & 146,643              & 3,378                \\
                        & TraDes~\cite{wu2021track}                   & 69.1                 & 63.9                 & 36.4                 & 21.5                 & 20,892               & 150,060              & 3,555                \\
                        & MAT~\cite{han2020mat}                      & 69.5                 & 63.1                 & 43.8                 & 18.9                 & 30,660               & 138,741              & 2,844                \\
                        & SOTMOT~\cite{zheng2021improving}                   & 71.0                 & 71.9                 & 42.7                 & 15.3                 & 39,537               & 118,983              & 5,184                \\
                        & RADTrack (RelationTrack)~\cite{yu2021relationtrack}                 & 73.1                 & 73.7                 & 39.9                 & 20.0                 & 25,935               & 122,700              & 3,021                \\
                        
                        & GSDT~\cite{wang2021joint}                   & 73.2                 & 66.5                 & 41.7                 & 17.5                 & 26,397               & 120,666              & 3,891                \\
                        & Semi-TCL~\cite{li2021semi}                 & 73.3                 & 73.2                 & 41.8                 & 18.7                 & 22,944               & 124,980              & 2,790                \\
                        & FairMOT~\cite{zhang2020fairmot}                  & 73.7                 & 72.3                 & 43.2                 & 17.3                 & 27,507               & 117,477              & 3,303                \\
                        & RelationTrack~\cite{yu2021relationtrack}            & 73.8                 & \textbf{74.7}        & 41.7                 & 23.2                 & 27,999               & 118,623              & \textbf{1,374}       \\
                        & PermaTrackPr~\cite{tokmakov2021learning}             & 73.8                 & 68.9                 & 43.8                 & 17.2                 & 28,998               & 115,104              & 3,699                \\
                        & CSTrack~\cite{liang2020rethinking}                  & \textbf{74.9}        & 72.6                 & 41.5                 & 17.5                 & 23,847               & \textbf{114,303}     & 3,567                \\
                        & \textbf{PatchTrack (ours)}        & 73.6                 & 65.2                 & \textbf{44.6}        & \textbf{12.5}        & 23,976               & 121,230              & 3,795                \\ \cline{2-9} \noalign{\vskip\doublerulesep
         \vskip-\arrayrulewidth} \cline{2-9}
                        & Transformer-based method  \\ \cline{2-9} 
                        & MOTR~\cite{zeng2021motr}                     & 65.1                 & \textbf{66.4}        & 33.0                 & 25.2                 & 45,486               & 149,307              & \textbf{2,049}       \\
                        & TrackFormer~\cite{meinhardt2021trackformer}              & 65.0                 & 63.9                 & 45.6                & 13.8                 & 70,443               & 123,552              & \textcolor{blue}{3,528}                \\
                        & MOTPrivate (TransCenter)~\cite{xu2021transcenter}              & 70.0                 & 62.1                 & 38.9                 & 20.4                & 28,119               & 136,722             & 4,647                \\
                        & TransCenter~\cite{xu2021transcenter}              & 73.2                 & 62.2                 & 40.8                 & 18.5                 & \textbf{23,112}               & 123,738              & 4,614                \\
                        & TrTrack (TransTrack)~\cite{sun2020transtrack}               & \textbf{75.2}        & 63.5                 & \textbf{55.3}                 & \textbf{10.2}        & 50,157               & \textbf{86,442}      & 3,603                \\
                        & \textbf{PatchTrack (ours)}        & \textcolor{blue}{73.6}                 & \textcolor{blue}{65.2}                 & \textcolor{blue}{44.6}        & \textcolor{blue}{12.5}                 & \textcolor{blue}{23,976}      & \textcolor{blue}{121,230}              & 3,795                \\ \hline
\end{tabularx}
\caption{\textbf{Evaluation on MOT17 test set.} We evaluate recent MOT systems on the MOT17 test set in a private detection protocol. Compared to CNN-based (non Transformer-based) methods, PatchTrack outperforms in MT and ML. We also compare our proposed method with MOT systems that are also Transformer based. Numbers are in bold if they are the best in their respective metric columns, and in blue if they are the second-to-best.}
\label{tab:experiments_mot17}
\end{table*}

\begin{figure*}
\centering
\begin{subfigure}[b]{0.3\textwidth}
\centering
\includegraphics[width=\textwidth]{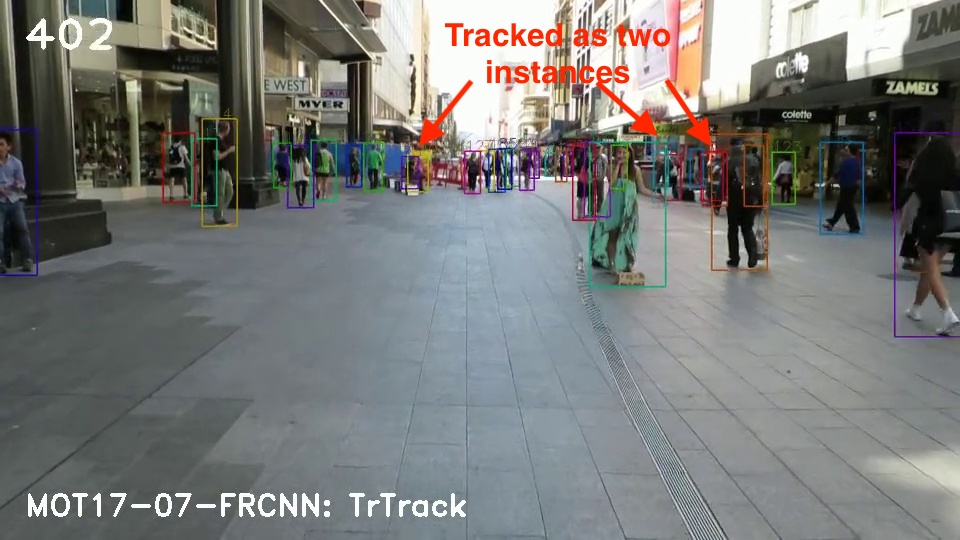}
\caption{TransTrack MOT17-07 Frame 402}
\label{fig:experiments_transtrack_mot17_07_402}
\end{subfigure}
\begin{subfigure}[b]{0.3\textwidth}
\centering
\includegraphics[width=\textwidth]{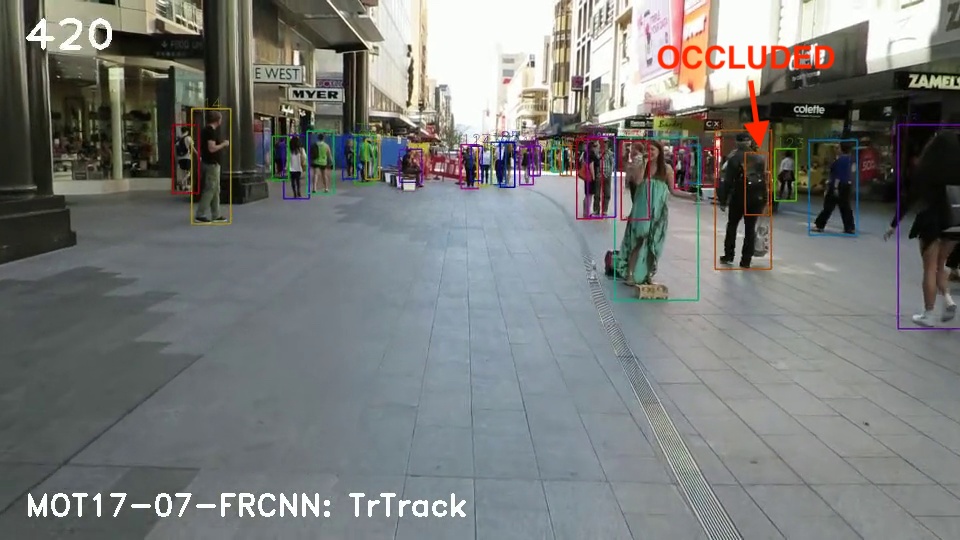}
\caption{TransTrack MOT17-07 Frame 420}
\label{fig:experiments_transtrack_mot17_07_420}
\end{subfigure}
\begin{subfigure}[b]{0.3\textwidth}
\centering
\includegraphics[width=\textwidth]{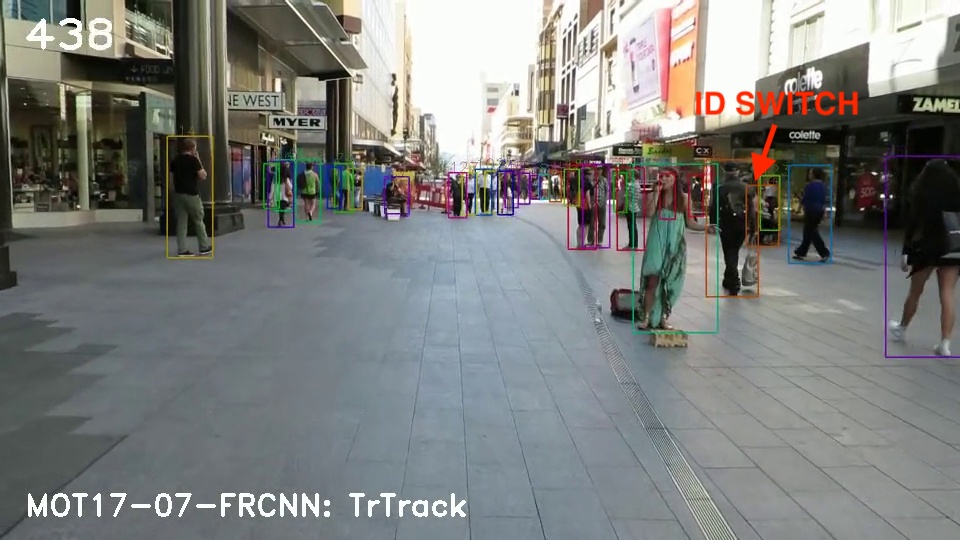}
\caption{TransTrack MOT17-07 Frame 438}
\label{fig:experiments_transtrack_mot17_07_438}
\end{subfigure}

\begin{subfigure}[b]{0.3\textwidth}
\centering
\includegraphics[width=\textwidth]{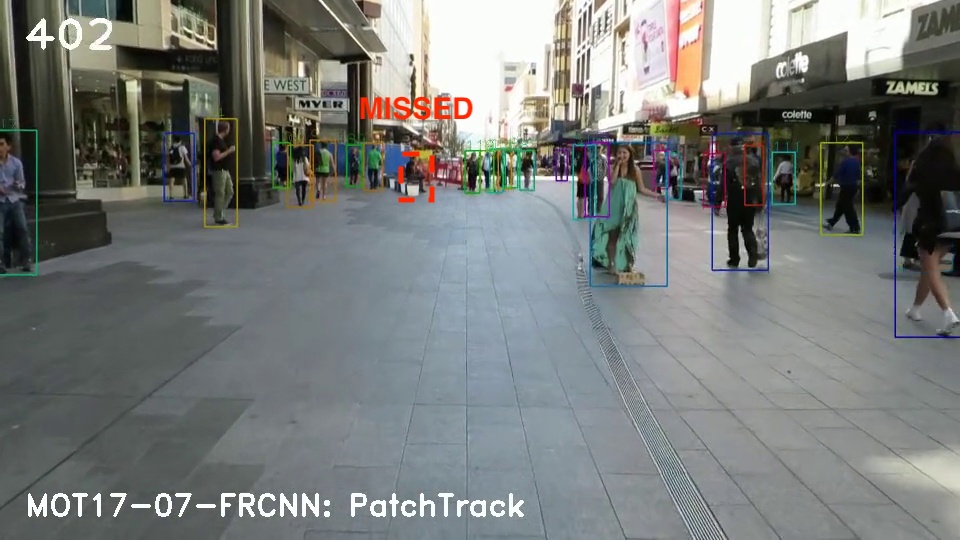}
\caption{PatchTrack MOT17-07 Frame 402}
\label{fig:experiments_patchtrack_mot17_07_402}
\end{subfigure}
\begin{subfigure}[b]{0.3\textwidth}
\centering
\includegraphics[width=\textwidth]{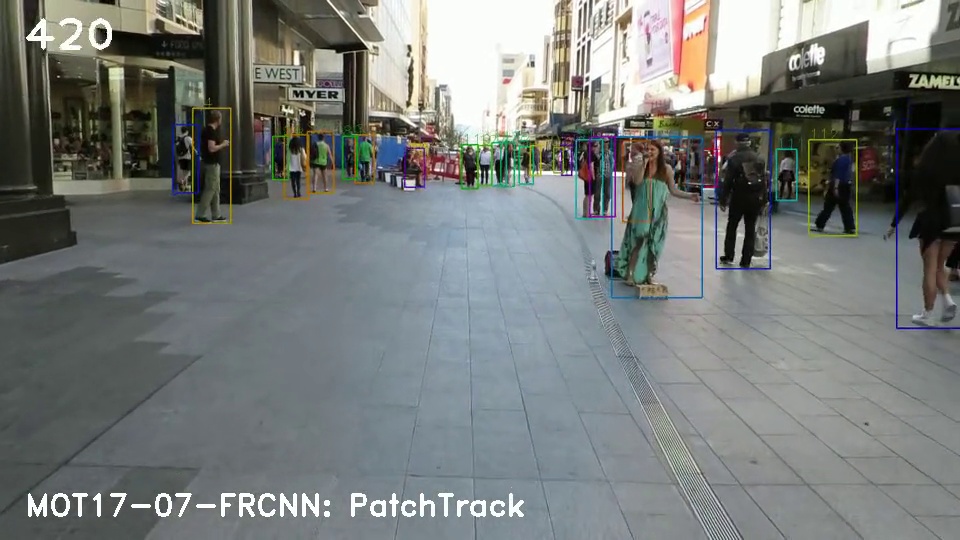}
\caption{PatchTrack MOT17-07 Frame 420}
\label{fig:experiments_patchtrack_mot17_07_420}
\end{subfigure}
\begin{subfigure}[b]{0.3\textwidth}
\centering
\includegraphics[width=\textwidth]{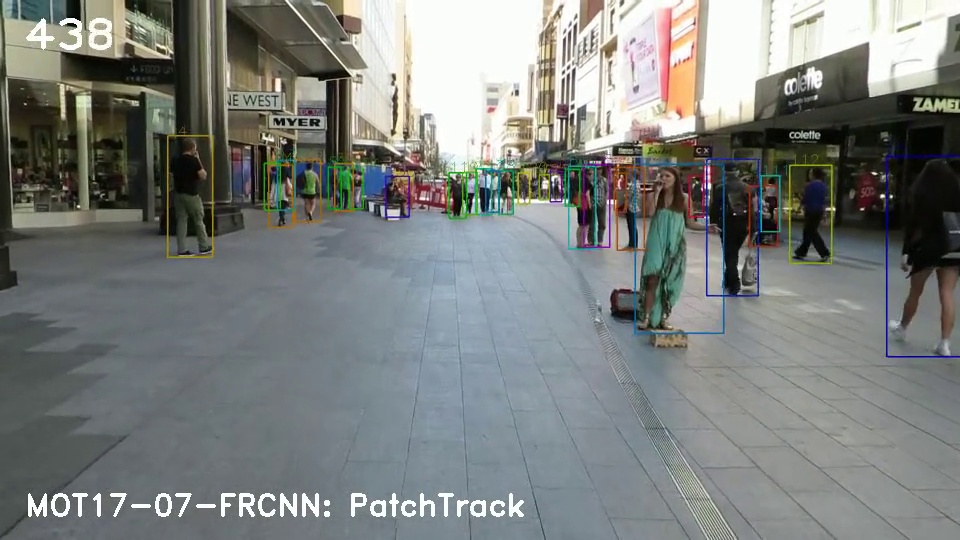}
\caption{PatchTrack MOT17-07 Frame 438}
\label{fig:experiments_patchtrack_mot17_07_438}
\end{subfigure}
\caption{\textbf{Visualizations on the MOT17 test set.} Comparing to TransTrack~\cite{sun2020transtrack}, PatchTrack is able to show comparable performance and generate less than 50\% FP, while TransTrack suffers from detecting one object multiple times (Figure~\ref{fig:experiments_transtrack_mot17_07_402}) and ID switches (Figure~\ref{fig:experiments_transtrack_mot17_07_438}) when trying to track fully-occluded objects (Figure~\ref{fig:experiments_transtrack_mot17_07_420}).}
\label{fig:experiments_mot17_visualization}
\end{figure*}

\section{Experiments}

\subsection{Datasets and metrics}

\noindent\textbf{MOT} MOT benchmarks are among the most widely used multi-object tracking benchmarks. We perform experiments on two of the MOT benchmarks: MOT16 and MOT17~\cite{milan2016mot16}. MOT16 consists of a training set of 7 videos (5,316 frames and 336,891 tracks) and a test set of 7 videos (5,919 frames and 564,228 tracks) with FPS ranging from 14 to 30. To evaluate the performance of the tracking mechanism independently of the detection accuracy, this benchmark also provides public detection from Faster R-CNN~\cite{ren2015faster}. MOT17 consists of the same training set and test set as MOT16, but with additional public detection from  DPM~\cite{felzenszwalb2009object} and SDP~\cite{yang2016exploit}. Both MOT16 and MOT17 are annotated with full-body bounding boxes. \vspace{0.1in}

\noindent\textbf{CrowdHuman} CrowdHuman~\cite{shao2018crowdhuman} is a pedestrian detection benchmark. It contains 15,000 training images and 4,370 validation images with a total of 470K objects. The annotations are also human full-body bounding boxes. This benchmark is often used for pre-training MOT systems. \vspace{0.1in}

\noindent\textbf{Metrics} MOT benchmarks~\cite{leal2015motchallenge, milan2016mot16, dendorfer2020mot20} uses metrics from CLEAR~\cite{bernardin2008evaluating}, which includes Multiple-Object Tracking Accuracy (MOTA), Identity F1 score (IDF1), Identity Switch (IDsw), False Positive (FP), False Negative (FN) detections, as well as Mostly Tracked (MT) and Mostly Lost (ML) trajectories.

\subsection{Training data generation} 

Given the architecture of PatchTrack (Figure~\ref{fig:method_patchtrack}), we need two consecutive frames to train the model. Although we could simply take frames pairs, predict track candidates from tracks of the previous frame using Kalman filter~\cite{welch1995introduction} as shown in the architecture, Kalman filter would not be able to provide high quality predictions due to high uncertainty in the early stage when there is a lack of prior information, which will in turn degrades the performance of decoder since the patch queries do not serve as good guesses to where existing tracks may be in the current frame. 

To simulate the role of Kalman filter~\cite{welch1995introduction} and generate track candidates for training, we propose the following augmentation strategy. Given a frame pair $f_{k-1}$ and $f_k$. We first randomly shift and reshape each track bounding box in frame $f_{k-1}$ within a pre-defined domain. We ensure that the IoU between each augmented bounding box and the track bounding box in frame $f_k$ with the same tracking ID, if exists, is at least 0.5. This is to align with commonly used IoU threshold value in detection association~\cite{Wojke2017simple, bewley2016simple, zhang2020fairmot}. These augmented tracks are the track candidates to our system during training. 

We also adapt the track augmentation strategy from Trackformer~\cite{meinhardt2021trackformer}, where we introduce false negatives by removing some queries associated with tracks that exist in both $f_{k-1}$ and $f_k$ from the input. The objective of the system is to detect the corresponding objects as new objects using object queries. On the other end, we sample output embeddings (generated from performing object detection on $f_{k-1}$) that map to background bounding boxes. They are included in the track queries as false positives when performing object tracking on $f_k$. To obtain their corresponding patch queries, we get their respective bounding boxes and augment them in the same manner as track candidates generation. We ensure that the IoU of each augmented bounding box is below 0.5 with ground truth tracks in $f_k$. For each patch-track queries generated from the above procedure, our system should decode them and get background objects. 

Frame pairs are selected from two sources. The first one is video data from MOT benchmarks~\cite{milan2016mot16}, where we take two video clips within a certain range from each other in the same video. This gives us more variety in terms of camera motion. The second one is image data from CrowdHuman~\cite{shao2018crowdhuman}, where we augment a single image through random scaling and translating to obtain a frame pair. For each selected frame pair, we perform the aforementioned steps to generate track candidates and modify the ground truth corresponding to false positives/negatives we inserted manually. PatchTrack is optimized towards the modified ground truth during training.

\subsection{Implementation details}

The Kalman filter~\cite{welch1995introduction} following a constant velocity model is used to predict track candidates. PatchTrack uses ResNet-50\cite{he2016deep} pre-trained on ImageNet~\cite{deng2009imagenet} as its CNN backbone and Deformable DETR~\cite{zhu2020deformable} for the Transformer encoder-decoder framework. The number of object queries is set to be 500. Inactive track queries will be kept for 30 frames for track re-birth. 

We adopt the training procedure from TransTrack~\cite{sun2020transtrack} as follows. The optimizer is AdamW with $\beta_1 = 0.9, \beta_2 = 0.999$ and initial learning rate $2\mathrm{e}{-4}$. We use 8 NVIDIA Tesla V100 GPUs with batch size 16. PatchTrack is first pre-trained on CrowdHuman~\cite{shao2018crowdhuman} for 150 epochs with the learning rate dropped to $2\mathrm{e}{-5}$ after the first 100 epochs. Then, PatchTrack is trained on both CrowdHuman and MOT17~\cite{milan2016mot16} for another 20 epochs. Lastly, it is evaluated on MOT16 and MOT17~\cite{milan2016mot16} test sets.

\subsection{Results}

\noindent\textbf{MOT16} We compare PatchTrack with other MOT systems on MOT16~\cite{milan2016mot16} test set in private protocol (Table~\ref{tab:experiments_mot16}), where PatchTrack achieves state-of-the-art results in MOTA, ML, and FN. Compared to LMP\_p~\cite{tang2017multiple} and POI~\cite{yu2016poi}, which collectively achieve best results in the remaining metrics, PatchTrack has significantly lower ML, showing overall better tracking performance. Figure~\ref{fig:experiments_mot16_visualization} shows additional visual comparison with LMP\_p and POI, where PatchTrack is able to track partially occluded objects and distinguish crowded objects better without missing objects or tracking one object multiple times.\vspace{0.1in}

\noindent\textbf{MOT17} Table~\ref{tab:experiments_mot17} shows quantitative results of PatchTrack along with other recent MOT systems on MOT17~\cite{milan2016mot16} test set in private protocol. Compared to Non-Transformer-based methods, PatchTrack reports best numbers in MT and ML, and shows superior ability in trajectory prediction. On the other hand, PatchTrack performs comparably well with other Transformer-based methods, achieving second-to-best results in most metrics. Compared to TransTrack~\cite{sun2020transtrack}, which has state-of-the-art results in MOTA, MT, ML, and FN, our system is able to produce less than 50\% of FP. We provide additional visualizations of PatchTrack and TransTrack in Figure~\ref{fig:experiments_mot17_visualization}. While PatchTrack is able to perform on par with TransTrack, our system is able to avoid tracking one object multiple times or causing ID switches when a previously fully occluded object re-appears. 

\subsection{Ablation study}

The ablation study is performed on the MOT17~\cite{milan2016mot16} validation set. The original MOT17 training set is split to a new training set and validation set, each consisting of the first half and the second half of training videos. After pre-training PatchTrack on CrowdHuman~\cite{shao2018crowdhuman}, the system is fine-tuned on the both CrowdHuman and the new MOT17 training set and evaluated on the validation set.\vspace{.1in}

\noindent\textbf{Type of queries} We evaluate the effect of various queries in Table~\ref{tab:experiments_ablation_query_type}. Removal of only patch queries or track queries means the other is sent to the Transformer decoder along with object queries. Removal of patch-track queries means that the decoder takes in object queries only and essentially behaves like an object detector. After getting individual detections for each frame, we use the Kalman filter~\cite{welch1995introduction} and the Hungarian algorithm~\cite{kenesei2002hungarian} to associate them. In this case, the modified system falls into the tracking-by-detection paradigm. We see that both patch queries and track queries play an important role in the joint-detection-and-tracking setting. On the other hand, the performance of the tracking-by-detection version of our system is overall comparable with PatchTrack, but produces more ID switches.\vspace{.1in} 

\begin{table}[]
\begin{tabular}{l|cccc}
\hline
Method                & MOTA & MT  & ML & IDsw \\ \hline
w/o patch queries       & 71.4 & 165 & 42 & 214  \\
w/o track queries       & 66.3 & 141 & 61 & 248  \\
w/o patch-track queries & 72.0 & 176 & 40 & 200  \\
PatchTrack            & 72.1 & 176 & 40 & 192  \\ \hline
\end{tabular}
\caption{\textbf{Ablation study on type of query inputs.} We send different types of query inputs to our system and evaluate their effects. The results suggest the positive effect of patch queries and track queries. When the system doesn't use patch-track queries and behave as an object detector, where we use Kalman filter~\cite{welch1995introduction} and Hungarian algorithm~\cite{kenesei2002hungarian} to associate predicted detections, the system produces more ID switches.}
\label{tab:experiments_ablation_query_type}
\end{table}

\noindent\textbf{Source of frame patches}
We also evaluate patch queries generated from different sources. The \textit{previous bboxes} patches come directly from cropping the current frame of interest using bounding boxes of tracks in the previous frame. Alternatively, the \textit{previous frame} patches are generated using both the previous frame and bounding boxes of tracks in the previous frame. From Table~\ref{tab:experiments_ablation_patch}, we see similar results when using patches from the previous frame compared to using track queries alone, meaning that patches from the previous frame contains similar information to track queries. On the other hand, patches generated from the current frame with bounding boxes of tracks in the previous frame degrade the performance. We reason that it is because of the misalignment between the frame and bounding boxes, which leads to less useful information in patches.

\begin{table}[]
\centering
\begin{tabular}{l|cccc}
\hline
Method          & MOTA & MT  & ML & IDsw \\ \hline
w/o patch query & 71.4 & 165 & 42 & 214  \\
previous bboxes & 62.8 & 137 & 69 & 258  \\
previous frame  & 71.4 & 165 & 42 & 214  \\
PatchTrack      & 72.1 & 176 & 40 & 192  \\ \hline
\end{tabular}
\caption{\textbf{Ablation study on source of frame patches.} We test patch queries generated from different sources. When the patches come from cropping the current frame using the track bounding boxes from the previous frame (\textit{previous bboxes}), the corresponding patch queries have a negative effect on the performance.}
\label{tab:experiments_ablation_patch}
\end{table}

%% file: sections/Conclusion.tex
\section{Conclusion}
We present PatchTrack, a Transformer-based joint-detection-and-tracking system using frame patches. By generating patch queries from the current frame of interest and track predictions using a motion model, we obtain information about object motion and appearance that is associated with the current frame. This novel way of using visual information in the current frame adds additional information to track queries that are derived from previous frames. By using both queries collectively, PatchTrack is able to achieve competitive results on MOT benchmarks.

%% file: arvix.bbl
\begin{thebibliography}{10}\itemsep=-1pt

\bibitem{babaee2019dual}
Maryam Babaee, Zimu Li, and Gerhard Rigoll.
\newblock A dual cnn--rnn for multiple people tracking.
\newblock {\em Neurocomputing}, 368:69--83, 2019.

\bibitem{bernardin2008evaluating}
Keni Bernardin and Rainer Stiefelhagen.
\newblock Evaluating multiple object tracking performance: the clear mot
  metrics.
\newblock {\em EURASIP Journal on Image and Video Processing}, 2008:1--10,
  2008.

\bibitem{bewley2016simple}
Alex Bewley, Zongyuan Ge, Lionel Ott, Fabio Ramos, and Ben Upcroft.
\newblock Simple online and realtime tracking.
\newblock In {\em 2016 IEEE international conference on image processing
  (ICIP)}, pages 3464--3468. IEEE, 2016.

\bibitem{cai2018cascade}
Zhaowei Cai and Nuno Vasconcelos.
\newblock Cascade r-cnn: Delving into high quality object detection.
\newblock In {\em Proceedings of the IEEE conference on computer vision and
  pattern recognition}, pages 6154--6162, 2018.

\bibitem{carion2020end}
Nicolas Carion, Francisco Massa, Gabriel Synnaeve, Nicolas Usunier, Alexander
  Kirillov, and Sergey Zagoruyko.
\newblock End-to-end object detection with transformers.
\newblock In {\em European Conference on Computer Vision}, pages 213--229.
  Springer, 2020.

\bibitem{dai2021up}
Zhigang Dai, Bolun Cai, Yugeng Lin, and Junying Chen.
\newblock Up-detr: Unsupervised pre-training for object detection with
  transformers.
\newblock In {\em Proceedings of the IEEE/CVF Conference on Computer Vision and
  Pattern Recognition}, pages 1601--1610, 2021.

\bibitem{dendorfer2020mot20}
Patrick Dendorfer, Hamid Rezatofighi, Anton Milan, Javen Shi, Daniel Cremers,
  Ian Reid, Stefan Roth, Konrad Schindler, and Laura Leal-Taix{\'e}.
\newblock Mot20: A benchmark for multi object tracking in crowded scenes.
\newblock {\em arXiv preprint arXiv:2003.09003}, 2020.

\bibitem{deng2009imagenet}
Jia Deng, Wei Dong, Richard Socher, Li-Jia Li, Kai Li, and Li Fei-Fei.
\newblock Imagenet: A large-scale hierarchical image database.
\newblock In {\em 2009 IEEE conference on computer vision and pattern
  recognition}, pages 248--255. Ieee, 2009.

\bibitem{fang2018recurrent}
Kuan Fang, Yu Xiang, Xiaocheng Li, and Silvio Savarese.
\newblock Recurrent autoregressive networks for online multi-object tracking.
\newblock In {\em 2018 IEEE Winter Conference on Applications of Computer
  Vision (WACV)}, pages 466--475. IEEE, 2018.

\bibitem{felzenszwalb2009object}
Pedro~F Felzenszwalb, Ross~B Girshick, David McAllester, and Deva Ramanan.
\newblock Object detection with discriminatively trained part-based models.
\newblock {\em IEEE transactions on pattern analysis and machine intelligence},
  32(9):1627--1645, 2009.

\bibitem{girshick2015fast}
Ross Girshick.
\newblock Fast r-cnn.
\newblock In {\em Proceedings of the IEEE international conference on computer
  vision}, pages 1440--1448, 2015.

\bibitem{girshick2014rich}
Ross Girshick, Jeff Donahue, Trevor Darrell, and Jitendra Malik.
\newblock Rich feature hierarchies for accurate object detection and semantic
  segmentation.
\newblock In {\em Proceedings of the IEEE conference on computer vision and
  pattern recognition}, pages 580--587, 2014.

\bibitem{Goodfellow-et-al-2016}
Ian Goodfellow, Yoshua Bengio, and Aaron Courville.
\newblock {\em Deep Learning}.
\newblock MIT Press, 2016.
\newblock \url{http://www.deeplearningbook.org}.

\bibitem{han2020mat}
Shoudong Han, Piao Huang, Hongwei Wang, En Yu, Donghaisheng Liu, Xiaofeng Pan,
  and Jun Zhao.
\newblock Mat: Motion-aware multi-object tracking.
\newblock {\em arXiv preprint arXiv:2009.04794}, 2020.

\bibitem{he2016deep}
Kaiming He, Xiangyu Zhang, Shaoqing Ren, and Jian Sun.
\newblock Deep residual learning for image recognition.
\newblock In {\em Proceedings of the IEEE conference on computer vision and
  pattern recognition}, pages 770--778, 2016.

\bibitem{kenesei2002hungarian}
Istv{\'a}n Kenesei, Robert~M Vago, and Anna Fenyvesi.
\newblock {\em Hungarian}.
\newblock Routledge, 2002.

\bibitem{leal2015motchallenge}
Laura Leal-Taix{\'e}, Anton Milan, Ian Reid, Stefan Roth, and Konrad Schindler.
\newblock Motchallenge 2015: Towards a benchmark for multi-target tracking.
\newblock {\em arXiv preprint arXiv:1504.01942}, 2015.

\bibitem{li2021semi}
Wei Li, Yuanjun Xiong, Shuo Yang, Mingze Xu, Yongxin Wang, and Wei Xia.
\newblock Semi-tcl: Semi-supervised track contrastive representation learning.
\newblock {\em arXiv preprint arXiv:2107.02396}, 2021.

\bibitem{liang2020rethinking}
Chao Liang, Zhipeng Zhang, Yi Lu, Xue Zhou, Bing Li, Xiyong Ye, and Jianxiao
  Zou.
\newblock Rethinking the competition between detection and reid in multi-object
  tracking.
\newblock {\em arXiv preprint arXiv:2010.12138}, 2020.

\bibitem{lin2017focal}
Tsung-Yi Lin, Priya Goyal, Ross Girshick, Kaiming He, and Piotr Doll{\'a}r.
\newblock Focal loss for dense object detection.
\newblock In {\em Proceedings of the IEEE international conference on computer
  vision}, pages 2980--2988, 2017.

\bibitem{lin2021detection}
Xufeng Lin, Chang-Tsun Li, Victor Sanchez, and Carsten Maple.
\newblock On the detection-to-track association for online multi-object
  tracking.
\newblock {\em Pattern Recognition Letters}, 146:200--207, 2021.

\bibitem{liu2016ssd}
Wei Liu, Dragomir Anguelov, Dumitru Erhan, Christian Szegedy, Scott Reed,
  Cheng-Yang Fu, and Alexander~C Berg.
\newblock Ssd: Single shot multibox detector.
\newblock In {\em European conference on computer vision}, pages 21--37.
  Springer, 2016.

\bibitem{mahmoudi2019multi}
Nima Mahmoudi, Seyed~Mohammad Ahadi, and Mohammad Rahmati.
\newblock Multi-target tracking using cnn-based features: Cnnmtt.
\newblock {\em Multimedia Tools and Applications}, 78(6):7077--7096, 2019.

\bibitem{meinhardt2021trackformer}
Tim Meinhardt, Alexander Kirillov, Laura Leal-Taixe, and Christoph
  Feichtenhofer.
\newblock Trackformer: Multi-object tracking with transformers.
\newblock {\em arXiv preprint arXiv:2101.02702}, 2021.

\bibitem{meng2021conditional}
Depu Meng, Xiaokang Chen, Zejia Fan, Gang Zeng, Houqiang Li, Yuhui Yuan, Lei
  Sun, and Jingdong Wang.
\newblock Conditional detr for fast training convergence.
\newblock In {\em Proceedings of the IEEE/CVF International Conference on
  Computer Vision}, pages 3651--3660, 2021.

\bibitem{milan2016mot16}
Anton Milan, Laura Leal-Taix{\'e}, Ian Reid, Stefan Roth, and Konrad Schindler.
\newblock Mot16: A benchmark for multi-object tracking.
\newblock {\em arXiv preprint arXiv:1603.00831}, 2016.

\bibitem{pang2020tubetk}
Bo Pang, Yizhuo Li, Yifan Zhang, Muchen Li, and Cewu Lu.
\newblock Tubetk: Adopting tubes to track multi-object in a one-step training
  model.
\newblock In {\em Proceedings of the IEEE/CVF Conference on Computer Vision and
  Pattern Recognition}, pages 6308--6318, 2020.

\bibitem{pang2021quasi}
Jiangmiao Pang, Linlu Qiu, Xia Li, Haofeng Chen, Qi Li, Trevor Darrell, and
  Fisher Yu.
\newblock Quasi-dense similarity learning for multiple object tracking.
\newblock In {\em Proceedings of the IEEE/CVF Conference on Computer Vision and
  Pattern Recognition}, pages 164--173, 2021.

\bibitem{peng2020chained}
Jinlong Peng, Changan Wang, Fangbin Wan, Yang Wu, Yabiao Wang, Ying Tai,
  Chengjie Wang, Jilin Li, Feiyue Huang, and Yanwei Fu.
\newblock Chained-tracker: Chaining paired attentive regression results for
  end-to-end joint multiple-object detection and tracking.
\newblock In {\em European Conference on Computer Vision}, pages 145--161.
  Springer, 2020.

\bibitem{redmon2016you}
Joseph Redmon, Santosh Divvala, Ross Girshick, and Ali Farhadi.
\newblock You only look once: Unified, real-time object detection.
\newblock In {\em Proceedings of the IEEE conference on computer vision and
  pattern recognition}, pages 779--788, 2016.

\bibitem{ren2015faster}
Shaoqing Ren, Kaiming He, Ross Girshick, and Jian Sun.
\newblock Faster r-cnn: Towards real-time object detection with region proposal
  networks.
\newblock {\em Advances in neural information processing systems}, 28:91--99,
  2015.

\bibitem{rezatofighi2019generalized}
Hamid Rezatofighi, Nathan Tsoi, JunYoung Gwak, Amir Sadeghian, Ian Reid, and
  Silvio Savarese.
\newblock Generalized intersection over union: A metric and a loss for bounding
  box regression.
\newblock In {\em Proceedings of the IEEE/CVF Conference on Computer Vision and
  Pattern Recognition}, pages 658--666, 2019.

\bibitem{shao2018crowdhuman}
Shuai Shao, Zijian Zhao, Boxun Li, Tete Xiao, Gang Yu, Xiangyu Zhang, and Jian
  Sun.
\newblock Crowdhuman: A benchmark for detecting human in a crowd.
\newblock {\em arXiv preprint arXiv:1805.00123}, 2018.

\bibitem{shuai2021siammot}
Bing Shuai, Andrew Berneshawi, Xinyu Li, Davide Modolo, and Joseph Tighe.
\newblock Siammot: Siamese multi-object tracking.
\newblock In {\em Proceedings of the IEEE/CVF Conference on Computer Vision and
  Pattern Recognition}, pages 12372--12382, 2021.

\bibitem{sun2020transtrack}
Peize Sun, Yi Jiang, Rufeng Zhang, Enze Xie, Jinkun Cao, Xinting Hu, Tao Kong,
  Zehuan Yuan, Changhu Wang, and Ping Luo.
\newblock Transtrack: Multiple-object tracking with transformer.
\newblock {\em arXiv preprint arXiv:2012.15460}, 2020.

\bibitem{sun2019deep}
ShiJie Sun, Naveed Akhtar, HuanSheng Song, Ajmal Mian, and Mubarak Shah.
\newblock Deep affinity network for multiple object tracking.
\newblock {\em IEEE transactions on pattern analysis and machine intelligence},
  43(1):104--119, 2019.

\bibitem{tang2017multiple}
Siyu Tang, Mykhaylo Andriluka, Bjoern Andres, and Bernt Schiele.
\newblock Multiple people tracking by lifted multicut and person
  re-identification.
\newblock In {\em Proceedings of the IEEE conference on computer vision and
  pattern recognition}, pages 3539--3548, 2017.

\bibitem{tokmakov2021learning}
Pavel Tokmakov, Jie Li, Wolfram Burgard, and Adrien Gaidon.
\newblock Learning to track with object permanence.
\newblock {\em arXiv preprint arXiv:2103.14258}, 2021.

\bibitem{vaswani2017attention}
Ashish Vaswani, Noam Shazeer, Niki Parmar, Jakob Uszkoreit, Llion Jones,
  Aidan~N Gomez, {\L}ukasz Kaiser, and Illia Polosukhin.
\newblock Attention is all you need.
\newblock In {\em Advances in neural information processing systems}, pages
  5998--6008, 2017.

\bibitem{veeramani2018deepsort}
Balaji Veeramani, John~W Raymond, and Pritam Chanda.
\newblock Deepsort: deep convolutional networks for sorting haploid maize
  seeds.
\newblock {\em BMC bioinformatics}, 19(9):1--9, 2018.

\bibitem{wan2018multi}
Xingyu Wan, Jinjun Wang, Zhifeng Kong, Qing Zhao, and Shunming Deng.
\newblock Multi-object tracking using online metric learning with long
  short-term memory.
\newblock In {\em 2018 25th IEEE International Conference on Image Processing
  (ICIP)}, pages 788--792. IEEE, 2018.

\bibitem{wang2021joint}
Yongxin Wang, Kris Kitani, and Xinshuo Weng.
\newblock Joint object detection and multi-object tracking with graph neural
  networks.
\newblock In {\em 2021 IEEE International Conference on Robotics and Automation
  (ICRA)}, pages 13708--13715. IEEE, 2021.

\bibitem{welch1995introduction}
Greg Welch, Gary Bishop, et~al.
\newblock An introduction to the kalman filter.
\newblock 1995.

\bibitem{Wojke2017simple}
Nicolai Wojke, Alex Bewley, and Dietrich Paulus.
\newblock Simple online and realtime tracking with a deep association metric.
\newblock In {\em 2017 IEEE International Conference on Image Processing
  (ICIP)}, pages 3645--3649. IEEE, 2017.

\bibitem{wu2021track}
Jialian Wu, Jiale Cao, Liangchen Song, Yu Wang, Ming Yang, and Junsong Yuan.
\newblock Track to detect and segment: An online multi-object tracker.
\newblock In {\em Proceedings of the IEEE/CVF Conference on Computer Vision and
  Pattern Recognition}, pages 12352--12361, 2021.

\bibitem{xu2021transcenter}
Yihong Xu, Yutong Ban, Guillaume Delorme, Chuang Gan, Daniela Rus, and Xavier
  Alameda-Pineda.
\newblock Transcenter: Transformers with dense queries for multiple-object
  tracking.
\newblock {\em arXiv preprint arXiv:2103.15145}, 2021.

\bibitem{yang2016exploit}
Fan Yang, Wongun Choi, and Yuanqing Lin.
\newblock Exploit all the layers: Fast and accurate cnn object detector with
  scale dependent pooling and cascaded rejection classifiers.
\newblock In {\em Proceedings of the IEEE conference on computer vision and
  pattern recognition}, pages 2129--2137, 2016.

\bibitem{yu2021relationtrack}
En Yu, Zhuoling Li, Shoudong Han, and Hongwei Wang.
\newblock Relationtrack: Relation-aware multiple object tracking with decoupled
  representation.
\newblock {\em arXiv preprint arXiv:2105.04322}, 2021.

\bibitem{yu2016poi}
Fengwei Yu, Wenbo Li, Quanquan Li, Yu Liu, Xiaohua Shi, and Junjie Yan.
\newblock Poi: Multiple object tracking with high performance detection and
  appearance feature.
\newblock In {\em European Conference on Computer Vision}, pages 36--42.
  Springer, 2016.

\bibitem{zeng2021motr}
Fangao Zeng, Bin Dong, Tiancai Wang, Cheng Chen, Xiangyu Zhang, and Yichen Wei.
\newblock Motr: End-to-end multiple-object tracking with transformer.
\newblock {\em arXiv preprint arXiv:2105.03247}, 2021.

\bibitem{zhang2020fairmot}
Yifu Zhang, Chunyu Wang, Xinggang Wang, Wenjun Zeng, and Wenyu Liu.
\newblock Fairmot: On the fairness of detection and re-identification in
  multiple object tracking.
\newblock {\em arXiv preprint arXiv:2004.01888}, 2020.

\bibitem{zheng2021improving}
Linyu Zheng, Ming Tang, Yingying Chen, Guibo Zhu, Jinqiao Wang, and Hanqing Lu.
\newblock Improving multiple object tracking with single object tracking.
\newblock In {\em Proceedings of the IEEE/CVF Conference on Computer Vision and
  Pattern Recognition}, pages 2453--2462, 2021.

\bibitem{zhou2020tracking}
Xingyi Zhou, Vladlen Koltun, and Philipp Kr{\"a}henb{\"u}hl.
\newblock Tracking objects as points.
\newblock In {\em European Conference on Computer Vision}, pages 474--490.
  Springer, 2020.

\bibitem{zhou2019objects}
Xingyi Zhou, Dequan Wang, and Philipp Kr{\"a}henb{\"u}hl.
\newblock Objects as points.
\newblock {\em arXiv preprint arXiv:1904.07850}, 2019.

\bibitem{zhou2018online}
Zongwei Zhou, Junliang Xing, Mengdan Zhang, and Weiming Hu.
\newblock Online multi-target tracking with tensor-based high-order graph
  matching.
\newblock In {\em 2018 24th International Conference on Pattern Recognition
  (ICPR)}, pages 1809--1814. IEEE, 2018.

\bibitem{zhu2020deformable}
Xizhou Zhu, Weijie Su, Lewei Lu, Bin Li, Xiaogang Wang, and Jifeng Dai.
\newblock Deformable detr: Deformable transformers for end-to-end object
  detection.
\newblock {\em arXiv preprint arXiv:2010.04159}, 2020.

\end{thebibliography}
